\documentclass[a4paper,UKenglish]{oasics}

\usepackage{microtype}

\usepackage{paralist}

\usepackage{xcolor}

\definecolor{sfbblue}{RGB}{0,97,161}
\definecolor{sfbred}{RGB}{226,0,32}
\definecolor{green3}{rgb}{0.1328,0.543,0.1328}

\usepackage{wrapfig}

\usepackage{natbib}

\newcommand{\clpqs}{CLP(QS)}

\title{{\LARGE Cognitive Interpretation of Everyday Activities}\newline{\normalsize --- Toward Perceptual Narrative Based Visuo-Spatial Scene Interpretation}}

\titlerunning{Narrative based Cognitive Interpretation of Everyday Activities} 

\author{\sffamily Mehul Bhatt, Jakob Suchan, Carl Schultz}
\affil{Cognitive Systems (CoSy), and Spatial Cognition Research Center (SFB/TR 8)\\
University of Bremen, Germany\\
  {\upshape\footnotesize\texttt{\color{blue!80!black}bhatt@informatik.uni-bremen.de}}}

\authorrunning{M. Bhatt et. al.}

\Copyright[by]{Mehul Bhatt}



\serieslogo{}
\volumeinfo
  {M. Finlayson., B. Fisseni., B. L\"{o}we., J. C. Meister}
  {2}
  {Computational Models of Narrative (CMN) 2013}
  {}
  {}
  {1}
\EventShortName{CMN 2013}

\DOI{tab}

\begin{document}

\maketitle

%

\begin{abstract}\small
We position a narrative-centred computational model for high-level knowledge representation and reasoning in the context of a range of assistive technologies concerned with \emph{visuo-spatial perception and cognition} tasks. Our proposed narrative model encompasses aspects such as \emph{space, events, actions, change, and interaction} from the viewpoint of commonsense reasoning and learning in large-scale cognitive systems. The broad focus of this paper is on the domain of \emph{human-activity interpretation} in smart environments, ambient intelligence etc. In the backdrop of a \emph{smart meeting cinematography} domain, we position the proposed narrative model, preliminary work on perceptual narrativisation, and the immediate outlook on constructing general-purpose open-source tools for perceptual narrativisation.

\end{abstract}



\section{Introduction: Cognitive Interpretation by Narrativisation}
Narratives have been a focus on study from several perspectives, most prominently from the viewpoint of language, literature, and computational linguistics; see for instance, discourse analysis and computational narratology \citep{Roland-1975-narrative-structural,CMN-Mani-2012,Mani-comp-narratology,Goguen-course-compu-narratology,FSS102323}. From the viewpoint of commonsense reasoning, and closely related to the computational models of narrative perspective, is the position of researchers in logics of \emph{action and change}; here, narratives are interpreted as ``\emph{a sequence of events about which we may have incomplete, conflicting or incorrect information}'' \citep{MillerS94,OccurNarraSC:Pinto:1998}. As per \citet{McCarthy:2000:concepts-logical-AI}, ``\emph{a narrative tells what happened, but any narrative can only tell a certain amount. A narrative will usually give facts about the future of a situation that are not just consequences of projection from an initial situation}''. The interpretation of narrative knowledge in this paper is based on these characterisations, especially in regard to the commonsense representation and reasoning tasks that accrue whilst modelling and reasoning about the perceptually grounded, narrativised epistemic state of an autonomous agent pertaining to \emph{space, actions, events, and change} \citep{Bhatt:RSAC:2012}. In particular, this encompasses a range of inference patterns such as: (a) spatio-temporal abduction for scenario and narrative completion \citep{Bhatt:STeDy:10}; (b) integrated inductive-abductive reasoning with narrative knowledge \citep{dubba-bhatt-2011}; (c) narrative-based postdiction for abnormality detection and planning \citep{CR-2013-Narra-CogRob}.



%
%
%

\textbf{Perceptual Narratives}.\quad These are declarative models of visual, auditory, haptic and other observations in the real world that are obtained via artificial sensors and / or human input. Declarative models of perceptual narratives can be used for interpretation and control tasks in the course of assistive technologies in everyday life and work scenarios, e.g., behaviour interpretation, robotic plan generation, semantic model generation from video, ambient intelligence and smart environments (e.g., see narrative based models in \citep{DBLP:journals/corr/abs-1202-3728,DBLP:conf/aaaiss/HajishirziM11,DBLP:journals/lalc/Mueller07,Bhatt:STeDy:10,dubba-bhatt-2011,CR-2013-Narra-CogRob}).

\textbf{High-Level Cognitive Interpretation and Control}.\quad Our research is especially concerned with large-scale cognitive interaction systems where high-level perceptual sense-making, planning, and control constitutes one of many AI sub-components guiding other low-level control and attention tasks. As an example, consider the \emph{smart meeting cinematography} domain in Listing 1. In this domain, \emph{perceptual narratives} as in Fig. 1 are generated based on perceived spatial change interpreted as interactions of humans in the environment. Such narratives explaining the ongoing activities are needed to anticipate changes in the environment, as well as to appropriately influence the real-time control of the camera system. 
To convey the meaning of the presentation and the speakers interactions with a projection, the camera has to capture the scene including the speakers gestures, slides, and the audience. 
E.g., in Fig. 1, when the speaker explains the slides, the camera has to capture the speaker and the corresponding information on the slides. To this end, the camera records an overview shot capturing the speaker and the projection, and zooms on the particular element when the speaker explains it in detail, to allow the viewer to follow the presentation and to get the necessary information. When the speaker continues the talk, the camera focuses on the speaker to omit unnecessary and distracting information. To capture reactions of the audience, e.g. comments, questions or applause, the camera records an overview of the attending people or close-up shots of the commenting or asking person.

\begin{figure}
\begin{center}
\colorbox{gray!18}{
\begin{minipage}{0.975\textwidth}
\begin{wrapfigure}{l}{3.1cm}
\centering

\includegraphics[width=0.19\textwidth]{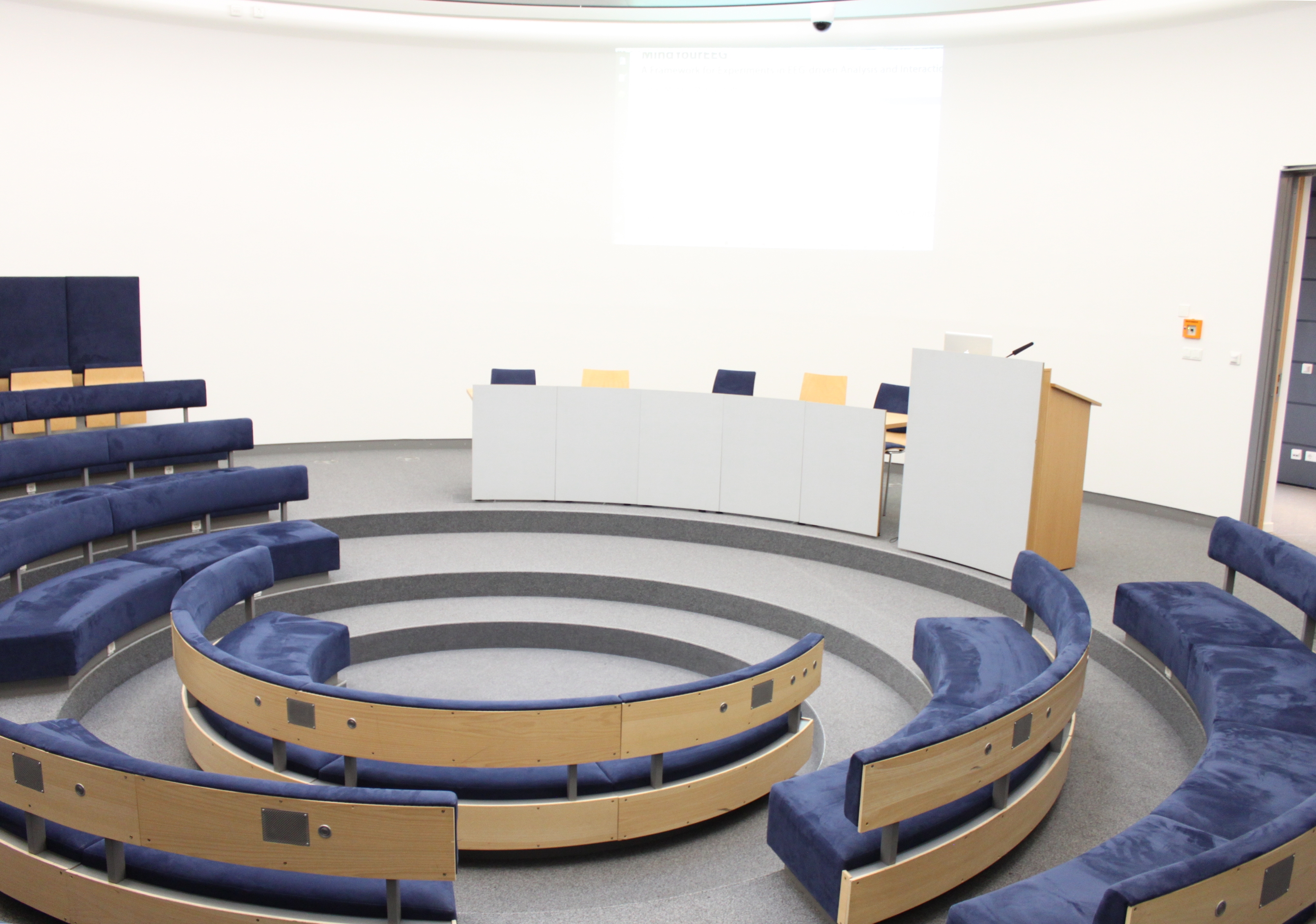}\\

\vspace{0.1in}

\hspace{0.04in}\includegraphics[width=0.19\textwidth]{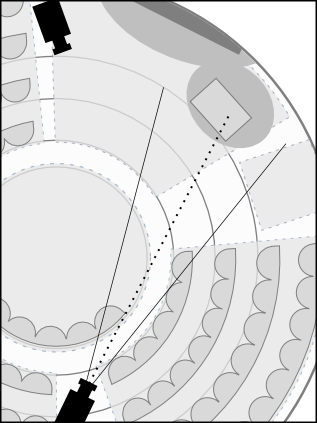}
\label{rc}
\end{wrapfigure}

\hfill{\footnotesize\textbf{Listing 1.\quad Smart Meeting Cinematography}

The smart meeting cinematography domain focusses on professional situations such as meetings and seminars. A basic task is to automatically produce dynamic recordings of interactive discussions, debates, presentations involving interacting people who use more than one communication modality such as hand-gestures (e.g., raising one's hand for a question, applause), voice and interruption, electronic apparatus (e.g., pressing of a button), movement (e.g., standing-up) and so forth.  The scenario consists of people-tracking, gesture identification closed under a context-specific taxonomy, and also involves real-time dynamic collaborative co-ordination and self-control of pan-tilt-zoom (PTZ) cameras in a \emph{sensing-planning-acting} loop. The long-term vision is to benchmark with respect to the capabilities of human-cinematographers, real-time video editors, surveillance personnel to record and semantically annotate individual and group activity (e.g., for summarisation, story-book format digital media and promo generation). \hfill\citep{Bhatt2013-Rotunde}}
\end{minipage}}
\end{center}

\vspace{-0.25in}

\end{figure}

\begin{figure}
\centering
\includegraphics[width=\textwidth]{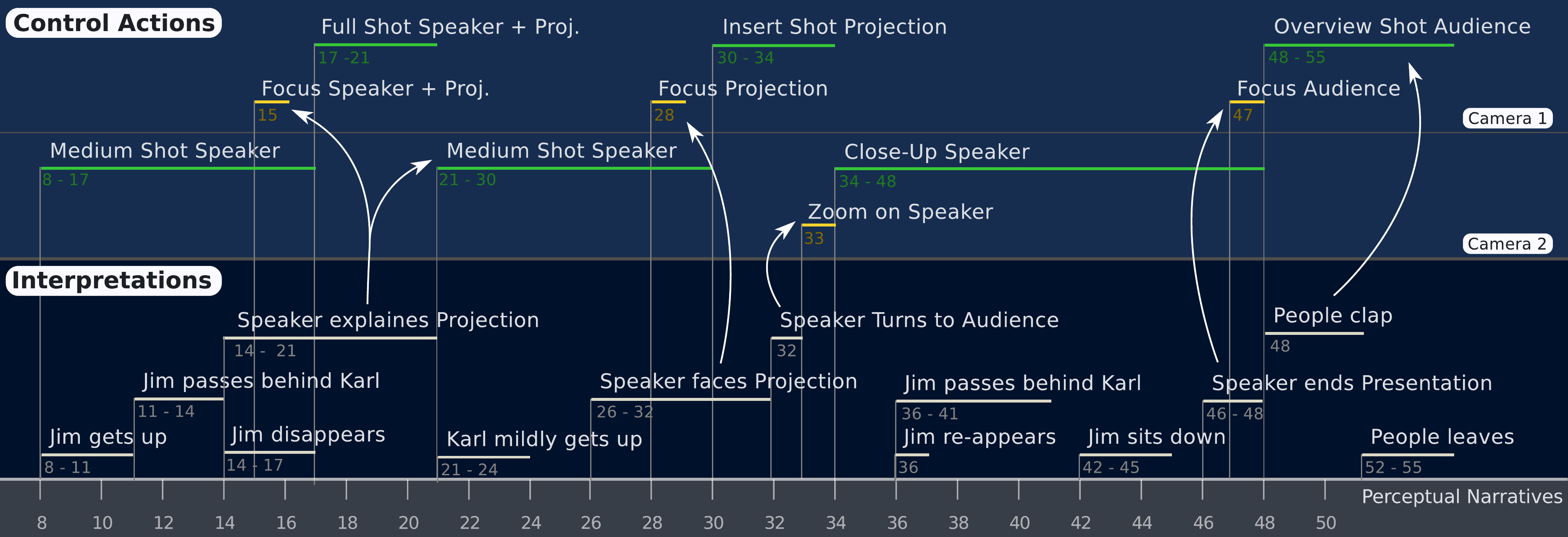}
\caption{Cognitive Interpretation and Control by Perceptual Narrativisation} 
\label{fig:rotunde-activities}
\end{figure}

\begin{figure}
\centering
\includegraphics[width=1\textwidth]{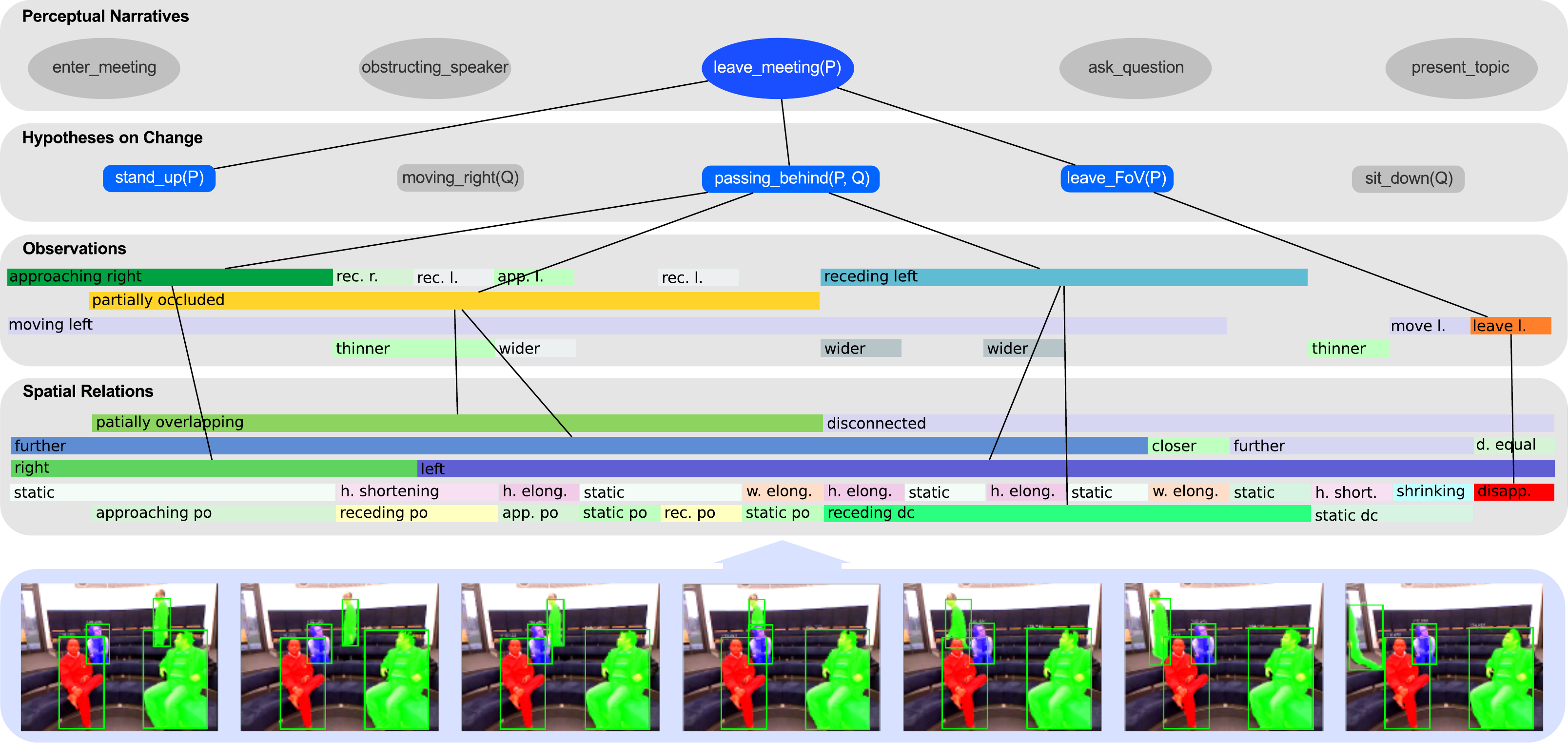}
\caption{Perceptual Narratives of Depth, Space, and Motion}
\label{pn}
\end{figure}


\newpage

\section{Perceptual Narrative Generation for Activity Interpretation}
Systems that monitor and interact with an environment populated by humans and other artefacts require a formal means for representing and reasoning about spatio-temporal, event and action based phenomena that are grounded to real public and private scenarios (e.g., logistical processes, activities of everyday living) of the environment being modelled.  A fundamental requirement within such application domains is the representation of \emph{dynamic} knowledge pertaining to the spatial aspects of the environment within which an agent, system, or robot is functional. This translates to the need to explicitly represent and reason about dynamic spatial configurations or scenes and, for real world problems, integrated reasoning about perceptual narratives of \emph{space, actions, and change} \citep{Bhatt:RSAC:2012}. With these modelling primitives, the ability to perform \emph{predictive} and \emph{explanatory} analyses on the basis of sensory data is crucial for creating a useful intelligent function within such environments \citep{dubba-bhatt-2011}.

\subsubsection*{Perceptual Narratives of Space and Motion}
To understand the nature of perceptual narratives (of space, and motion), consider the aforediscussed work-in-progress domain of \emph{smart meeting cinematography} (Listing 1). The particular infrastructural setup for the example presented herein consists of Pan-Tilt-Zoom (PTZ) capable cameras, depth sensors (Kinect), and a low-level vision module for people tracking (whole body, hand gesture, movement) customised on the basis of open-source algorithms and software. With respect to this setup, declaratively grounded perceptual narratives capturing the information in Fig. \ref{fig:rotunde-activities} is developed on the basis of a commonsense theory of \emph{qualitative space} (Listing 2), and interpretation of motion as qualitative spatial change \citep{Galton-2000-qspat-change}. In particular, the overall model as depicted in Fig. \ref{pn} consists of:

\smallskip
\emph{Space and Motion}: A theory to declaratively reason about qualitative spatial relations (e.g., topology, orientation), and qualitative motion perceived in the environment and interpret changes as domain dependent observations in the context of everyday activities involving humans and artefacts.

\smallskip
\emph{Explanation of (Spatial) Change}: Hypothesising real-world (inter)actions of individuals explaining the observations by integrating the qualitative theory with a learning method (e.g., Bayesian and Markov based (logic) learning) to incorporate uncertainty in the interpretation of observation sequences.

\smallskip
\emph{Semantic characterisation}: as a result of the aforementioned, real-time generation of declarative narratives of perceptual data (e.g., RGB-D) obtained directly from people/object tracking algorithms.
	
\smallskip

\begin{figure}
\begin{center}
\colorbox{gray!18}{
\begin{minipage}{0.98\textwidth}


\hfill{\footnotesize\textbf{Listing 2.\quad Qualitative Abstractions of Space and Motion } 

To represent space and spatial change we consider spatio-temporal relations \citep{CohnRenz2007} holding between individuals in the environment, i.e. \emph{topology, orientation, size, movement}. Combinations of spatial and temporal relations serve as observations describing perceived phenomena in the real world.  
\newline
The theory is implemented using \textbf{\clpqs} \citep{bhatt-et-al-2011}, which is a \emph{declarative spatial reasoning framework} that can be used for representing and reasoning about high-level, qualitative spatial knowledge about the world. CLP(QS) implements the semantics of qualitative spatial calculi within a constraint logic programming framework (amongst other things, this makes it possible to use spatial entities and relations between them as native entities). Furthermore it provides a declarative interface to qualitative and geometric spatial representation and reasoning capabilities such that these may be integrated with general knowledge representation and reasoning (KR) frameworks in artificial intelligence.}

\end{minipage}}
\end{center}
\vspace{-0.27in}

\end{figure}






Hypothesised object relations can be seen as building blocks to form complex interactions that are semantically interpreted as activities in the context of the domain. 
As an example consider the sequence of observations in the meeting environment depicted in Fig. \ref{pn}.

\begin{quote}
{\sffamily\small Region P {\color{blue} elongates vertically}, region P {\color{blue} approaches} region Q from the {\color{blue} right}, region P {\color{blue} partially overlaps} with region Q while P being {\color{blue} further away} from the observer than Q, region P {\color{blue} moves left}, region P {\color{blue} recedes} from region Q at the {\color{blue} left}, region P gets {\color{blue} disconnected} from region Q, region P {\color{blue} disappears} at the left border of the field of view} 
\end{quote}

To explain these observations in the `context' of the meeting situation we make hypothesis about possible interactions in the real world.

 \begin{quote}
{\sffamily\small Person P {\color{blue} stands up}, {\color{blue} passes behind} person Q while {\color{blue} moving towards} the exit and {\color{blue} leaves} the room. 
}
\end{quote}

The \textbf{semantic interpretation of activities} from video, depth (e.g., time-of-flight devices such as Kinect), and other forms of sensory input requires the representational and inferential mediation of qualitative abstractions of space, action, and change. Such relational abstractions serve as a bridge between high-level domain-specific conceptual or activity theoretic knowledge, and low-level statistically acquired features and sensory grammar learning techniques. Generation of perceptual narratives, and their access via the declarative interface of logic programming facilitates the integration of the overall framework in bigger projects concerned with cognitive vision, robotics, hybrid-intelligent systems etc. 
In the smart meeting cinematography domain the generated narratives are used to explain and understand the observations in the environment and anticipate interactions in it to allow for intelligent coordination and control of the involved PTZ-cameras.



\section{Immediate Outlook}
The smart meeting cinematography scenario presented in this paper serves as a challenging benchmark \citep{Bhatt2013-Rotunde} to investigate narrative based high-level cognitive interpretation of everyday interactions. Work is in progress to release certain aspects (pertaining to space, motion, real-time high-level control) emanating from the narrative model via the interface of constraint logic programming (e.g., as a Prolog based library of depth--space--motion). We also plan to release general tools to perform management and visualisation of activity interpretation data.

\newpage


\bibliographystyle{abbrvnat}


\end{document}